\pdfoutput=1

\documentclass[11pt]{article}

\usepackage[]{emnlp2021}

\usepackage{times}
\usepackage{latexsym}

\usepackage[T1]{fontenc}

\usepackage[utf8]{inputenc}

\usepackage{microtype}

\usepackage{graphicx}
\usepackage{comment}
\usepackage{booktabs}
\usepackage{url}
\usepackage{enumitem}
\usepackage{float}
\usepackage{xspace}
\usepackage{multirow}

\usepackage{amssymb}
\usepackage{pifont}
\newcommand{\xmark}{\ding{55}}%

\newcommand{\turkcorpus}{\textsc{TurkCorpus}\xspace}
\newcommand{\hsplit}{\textsc{HSplit}\xspace}
\newcommand{\asset}{\textsc{ASSET}\xspace}

\newcommand{\questeval}{\mbox{\textsc{QuestEval}}\xspace}

\title{Rethinking Automatic Evaluation in Sentence Simplification}

\author{Thomas Scialom$^{1,4,\thanks{~~Equal Contribution}}$~,  Louis Martin$^{2,3,\footnotemark[1]} $~,
\\{\bf Jacopo Staiano$^{4}$, Éric Villemonte de la Clergerie$^{3}$, Benoît Sagot$^3$ } \\
$^1$ Sorbonne Universit\'e, CNRS, LIP6, F-75005 Paris, France\\
$^2$ Facebook AI Research, Paris, France \\
$^3$ Inria, Paris, France \\
$^4$ reciTAL, Paris, France \\
  {\tt firstname@recital.ai, firstname.lastname@inria.fr}} 

\begin{document}
\maketitle

\begin{abstract}

In the context of Sentence Simplification, automatic evaluation is particularly challenging: by definition, the task requires to replace complex expressions with simpler ones that share the same meaning, limiting the effectiveness of n-gram based metrics like BLEU. 
Going hand in hand with the recent advances in NLG, new metrics have been proposed, such as BERTScore for Machine Translation. In summarization, the \questeval metric proposes to automatically compare two texts by questioning them.

In this paper, we first propose a simple adaptation of \questeval to Sentence Simplification tasks. 
We then extensively evaluate the correlations w.r.t.~human judgement for several metrics and show that BERTScore and \questeval obtains the best results. More importantly, we show that a large part of the correlations are actually spurious. To investigate this phenomenon further, we release a new corpus of evaluated simplifications, allowing us to remove the spurious correlations. We draw very different conclusions from the original ones, leading to a better understanding of the evaluated metrics. In particular, we raise concerns about very low correlations for most of the metrics: the only significant measure for Meaning Preservation comes from the proposed adaptation of \questeval.

\end{abstract}

\section{Introduction}

Sentence Simplification is a Natural Language Generation (NLG) task aiming at rewriting a text to make it easier to read and understand by using simpler words and syntax, while preserving its original meaning.
Sentence Simplification can benefit non-native speakers \cite{paetzold-specia:2016:AAAI}, as well as people with cognitive disabilities such as aphasia \cite{carroll-etal:1998}, dyslexia \cite{rello2013simplify} and autism \cite{evans-etal:2014}.
The literature of the field is prolific, with authors regularly claiming the state-of-the-art for their systems \cite{xu-etal-2016-optimizing,zhao-etal:2018:dmass-dcss,martin-etal:2020:access}.
However, automatic evaluation for NLG is known to be an open research question \cite{peyrard:2019:metrics-summarisation, scialom2020discriminative}, and Sentence Simplification is no exception \cite{xu-etal-2016-optimizing,sulem-etal:2018:hsplit,alva2020asset}.
On one hand, three main dimensions are usually evaluated: (i)~Fluency, (ii)~Simplicity, and (iii)~Meaning preservation. On the other hand, two metrics are widely used to compare systems: (i)~BLEU \cite{papineni-etal:2002:Bleu} and (ii)~SARI \cite{xu-etal-2016-optimizing}. 
BLEU is reported to better correlate for Meaning Preservation, while SARI for Simplicity (see 4.4 in \cite{xu-etal-2016-optimizing}). However, they have subsequently been shown to have low correlation with human judgements for various settings of Sentence Simplification \cite{sulem-etal:2018:hsplit,alva2020asset}.

Recent works, building on progress in Language Modeling, have shown promising results by computing pairwise token-level similarity of BERT representations \cite{Zhang-etal:2020:bertscore}. 
However, all these metrics operate at the token level, including BERTScore.
Token-level metrics have important theoretical limitations \cite{novikova-etal:2017:need-new-metrics-nlg}. In particular, not enough human references are available to cover all  possible ways to write the same idea. 

Beyond token-level metrics, \questeval\cite{scialom2021safeval} has recently obtained promising results in measuring Meaning Preservation in the summarization task. However, it uses exact matches between tokens to compute an F1 score, penalizing the use of synonyms and reformulations, hence preventing a direct application for Sentence Simplification.
In this paper, we propose a simple modification of \questeval to adapt it to Sentence Simplification. 

Further, to the best of our knowledge, no work has yet study the correlations for any of these recent metrics.
We show that both BERTScore and \questeval help in measuring human judgement for Sentence Simplification. They largely improve over BLEU and SARI, achieving new state-of-the-art correlations on all measured dimensions: Fluency, Meaning Preservation and Simplicity. While this result is not surprising for Meaning Preservation, it is rather unexpected for Simplicity: neither BERTScore or \questeval should have the ability to measure the simplicity of a text. Indeed, \questeval only compares the factual content of two texts, irrespective of their complexity. For its part, BERTScore is robust to synonyms and sentence structure; while this behavior is desired in Machine Translation, this is not the case in Sentence Simplification where a simpler word or sentence structure should be scored higher.

Our results indicate the presence of spurious correlations. We hypothesise that the inter-correlations between the evaluated dimensions could be responsible for this phenomenon: a system that generates simplifications that are not fluent tends to perform poorly also on Meaning Preservation and Simplicity.

Moreover, we are coming to a point where the outputs from neural systems are now close to a human-level for their Fluency \cite{zellers2019defending, scialom2020coldgans,martin2020multilingual}. We hypothesise that under this state of Fluency, the correlations of automatic metrics might vanish w.r.t.~human judgement. 

To investigate such phenomenon, we propose to analyse the correlations on human written simplifications: such texts should be less prompt to spurious correlations given that most of them should be fluent. To this purpose, we release a new human evaluation of \emph{human written} simplifications.\footnote{\url{http://dl.fbaipublicfiles.com/questeval/simplification_human_evaluations.tar.gz}}
This corpus allows us to conduct extensive experiments and better analyse the metrics' correlations w.r.t.~human judgement. 
In particular, our findings show very different conclusions than the evaluation of system-generated simplification. For instance, neither BLEU or SARI are significantly correlating with any dimensions. 

In summary, our contributions are:
\begin{enumerate}
    \item  We propose an adaptation of \questeval for Sentence Simplification and show that it compares favorably on Meaning Preservation. 
    \item We release a new large corpus of human evaluation. 
    \item  We conduct an extensive analysis of several metrics, including for the first time the recent BERTScore and our adaptation of \questeval. We study their correlations on both system-generated and human-written simplifications and show that most of correlations with Simplicity and Meaning Preservation are in fact spurious. On human-written simplifications, the only metric that actually correlates significantly for Meaning Preservation is \questeval.
\end{enumerate}

\section{Related Work}
\label{related_work}

\paragraph{Sentence Simplification -- Datasets}
BLEU and SARI both require ground truth references for proper evaluation.
For more accurate scores, these metrics require multiple references to account for all correct variations of simplifications.
To this end, \citet{xu-etal:2016} introduced \turkcorpus, an evaluation benchmark composed of 2,359 source sentences each associated with 8 human-written simplifications.
\turkcorpus however mostly focuses on lexical simplification and very light editing of the source sentence.
As a solution, \citet{alva2020asset} proposed \asset, a simplification dataset where references feature diverse rewriting operations, thus being able to evaluate a wider range of simplification models.
\asset reuses the same 2,359 source sentences as \turkcorpus, and provides 10 references per source.

\paragraph{Sentence Simplification -- Human Evaluation}: 
However, as automatic metrics serve as a proxy for human judgements, good metrics would be expected to have good correlations with human ratings.

\cite{xu-etal-2016-optimizing} studied the correlation between automatic metrics and human judgements and found significant correlations for SARI with Fluency, Meaning, and Simplicity, and for BLEU with Fluency and Meaning but not with Simplicity.
Human judges rated simplifications that were either system-generated or human-written.
Given poor system performance at the time, significant correlations could be due to the very different quality of simplifications between systems and humans only, but not be able to differentiate different systems.
In this work, we propose to distinguish correlations with system-generated simplifications and with human-written simplifications.

In \cite{alva2020asset}, the authors released a corpus of a total of 9,000 ratings of system-generated simplifications generated on their proposed dataset \asset and scored by human annotators on a Likert Scale.
They observed lower correlations of automatic metrics with human judgments than previously reported.
To the best of our knowledge this is so far the largest published human rating dataset of \emph{system-generated simplifications} for Sentence Simplification.
In this work, we propose to complement this dataset with 9000 ratings of \emph{human written} simplifications and show that we obtain different conclusions.

\begin{table*}[]
\centering\small
\begin{tabular}{l|l|l|c|c}
\toprule
\multicolumn{5}{l}{\begin{tabular}[c]{@{}l@{}} \textbf{Source Text:} In the Soviet years, the Bolsheviks demolished two of Rostov's principal landmarks- \\ St Alexander Nevsky  cathedral (1908) and St George cathedral in Nakhichevan (1783-1807).\end{tabular}} \\
\midrule
\multicolumn{5}{l}{\begin{tabular}[c]{@{}l@{}}\textbf{Simplification:} The Bolsheviks destroyed St. Alexander Nevsky cathedral and St. George cathedral in \\ Nakhichevan during the Soviet years.\end{tabular}}                                                  \\
\midrule[\heavyrulewidth]
\multicolumn{1}{c}{\multirow{2}{*}{\textbf{Generated Question}}}            & \multicolumn{2}{c}{\textbf{Answers}}                                                        & \multicolumn{2}{c}{\textbf{Score}}                                           \\
\multicolumn{1}{c}{}                                                        & \multicolumn{1}{c}{On Source}            & \multicolumn{1}{c}{On Simplif.}            & \multicolumn{1}{c}{F1}            & \multicolumn{1}{c}{BERTScore}            \\
\midrule
When did the Bolsheviks demolish St George cathedral?                       & the Soviet years                         & Soviet years                                     & 0.8                               & 0.89                                     \\
Who demolished St Alexander Nevsky cathedral?                               & demolished                               & destroyed                                        & 0.0                               & 0.82                                     \\
How many of Rostov's main landmarks were demolished?                        & two                                      & Unanswerable                                     & 0.0                               & 0.0                             \\
What cathedral was demolished in 1908?                                      & Rostov                                   &  Unanswerable                                    & 0.0                               & 0.0                             \\
{[}...{]}                                                                   & {[}...{]}                                & {[}...{]}                                        & {[}...{]}                         & {[}...{]}     \\                          
\bottomrule
\end{tabular}
\caption{Example of questions automatically generated and answered by \questeval given a source text and its simplification. }
\label{table:example_questeval_questions}
\end{table*}

\section{Human Evaluation Corpora}
\label{sec:dataset}

In this section we describe the two human ratings corpora used to compute the metrics correlations: they provide assessments over simplifications provided by automatic systems and humans. 

\paragraph{System-Likert}
We reuse the existing human evaluation corpus described in Section~\ref{related_work} on \asset. It is composed of ratings on systems-generated simplifications on a Likert Scale.
Each simplifications have been evaluated over three dimensions:
\begin{enumerate}
\itemsep0em 
    \item \textbf{Fluency}: how fluent is the evaluated text?
    \item \textbf{Meaning Preservation}: how well the evaluated text expresses the original meaning?
    \item \textbf{Simplicity}: to what extent is the evaluated text easier to read and understand?
\end{enumerate}
In total, 100 unique simplifications were evaluated with, for each of them, 30 ratings per aspect and per sentence.

\paragraph{Human-Likert}
We collect this second corpus following the same methodology used for \emph{System-Likert}, obtaining 9000 ratings of \emph{human written} simplifications sampled from the references available in  the test sets of \asset and \turkcorpus, and scored by human annotators given a 5-point Likert scale (1: Very Low, 5: Very High).
Details on the collection methodology are presented in the Appendix (Section~\ref{collection_methodology}).

\section{Metrics considered}

\subsection{Token-Level Metrics}

\paragraph{FKGL} \cite{kincaid-etal:1975:FK} is a reference-less metric that measures readability using only sentence lengths and word lengths.

\paragraph{SARI} \cite{xu-etal-2016-optimizing} was designed for Sentence Simplification by explicitly measuring the accuracy and recall of words that are added, deleted and kept.

\paragraph{BLEU} \cite{papineni-etal:2002:Bleu} measures the overlap of n-grams between a reference text and the evaluated one.

\paragraph{BERTScore} \cite{Zhang-etal:2020:bertscore} has been proposed for Machine Translation, leveraging on the contextualised representation of BERT to compute the similarity between the tokens. 

We note that these token-level metrics share the same limitation: the metric depends on the number of available references; given a small number of references, their correlations w.r.t. human judgement will decrease.

\subsection{\questeval for Sentence Simplification}

Moving beyond token-level metrics, a trend of using QG / QA for Automatic Summarization evaluation has recently emerged \questeval \cite{chen2017semantic, scialom2019answers, scialom2021safeval}.

\paragraph{\questeval} evaluates if a summary is factually consistent w.r.t.~its source document. To do so, it (i)~generates a list of questions on the evaluated summary, and (ii)~retrieves the corresponding answers from the source document: if the answers are similar, the summary is deemed satisfactory.\footnote{The \questeval metric is depicted in more detail in Figure~1 of the original paper \cite{scialom2021safeval}.}

\paragraph{Adapting \questeval to Sentence Simplification} To measure the similarity between two answers, the most popular approach in Question Answering is to compute the F1 score, as popularized by the SQuAD paper \cite{rajpurkar2016squad}, hence simply computing the 1-gram overlap between the retrieved answer and the gold reference for each generated question.
This is effective in the context of extractive Question Answering, since the answer belongs by definition to the input paragraph. In \questeval, the authors chose to compute the similarity via this F1. It may be considered an appropriate choice for summarization since extractive systems are reported to be very competitive baselines (e.g.~LEAD3 and TextRank), indicating that summarization does not require much rephrasing on the standard datasets. 

However, for Sentence Simplification, using synonyms and reformulations is inherent to the task. To alleviate this limitation, we propose to replace the F1 score with a more suitable metric: BERTScore. By leveraging its dense representations, a smoother similarity function than the F1 can be computed, allowing to take synonyms into account.

In Table~\ref{table:example_questeval_questions} we show an example of a source text, its simplification and some of the questions generated by \questeval. We observe that the simplification used a synonym, replacing \emph{demolished} with \emph{destroyed}. When a question was asked about this fact (see the second question in the Table), the QA model answered either \emph{destroyed} or \emph{demolished} w.r.t.~the given texts. For this example, the F1 Score incorrectly penalizes this answer, as opposed to BERTScore which is able to deal with synonyms.\footnote{It is also interesting that the third question (\emph{How many of Rostov’s main landmarks were demolished?}) was predicted to be unanswered. While the answer, i.e.~\emph{two}, could be deduced from the text, it could not extracted. This emphasizes a current limitation for \questeval, which could largely benefit from better and more abstractive QA models in the future.}

\begin{table*}[]
\centering\small
\begin{tabular}{lccrrrcrrr}\toprule
                & {\bf Ref-less}       && \multicolumn{3}{c}{\bf System-generated simplifications}     && \multicolumn{3}{c}{\bf Human-written simplifications}    \\
                &        && Fluency       & Simplicity      & Meaning     && Fluency       & Simplicity      & Meaning    \\
\midrule
Fluency         &              && ---         & 86.2**          & 79.5**  && ---         & 73.6**          & 52.7**          \\
Simplicity      &              && 86.2**          & ---         & 67.2**  && 73.6**          & ---         & 37.0**           \\
Meaning         &              && 79.5**          & 67.2**          & --- && 52.7**          & 37.0**          & ---         \\
\midrule
FKGL           & \checkmark   && -16.8           & -8.9            & -28.9*  && \textbf{19.0}            & \textbf{34.7}*  & 2.9             \\
SARI            & \xmark       && 18.3            & 25.2            & 16.0    && 0.9             & 9.7             & 5.8           \\
BLEU            & \xmark       && 37.9**          & 30.2*           & 41.1**  && 15.2            & 12.1            & 9.8             \\
BERTScore       & \xmark       && \textbf{53.6}** & \textbf{41.5}** & 63.3**  && 13.8            & 8.7             & 19.4            \\
\questeval        & \checkmark   && 45.8**          & 37.3**          & \textbf{66.5}** && -7.5            & -7.4   & \textbf{21.7}*         \\\bottomrule
\end{tabular}
\caption{Pearson Correlation Coefficient between human judgement and automatic metrics for system-generated simplification (left-hand), and for human-generated simplifications (right-hand). We report -FKGL so higher is better for all the metrics (a lower FKGL is supposed to indicate a simpler text). *~indicates p-value $<$~0.01 and **~$<$~0.001.}
\label{table:correlations}
\end{table*}

\paragraph{To BERTScore or not to BERTScore?}

Like BLEU, BERTScore is a \emph{token-level metric}, and therefore suffers from token misalignment: two texts can share the same meaning but be written in very different ways. The longer the texts, the more likely their tokens won't be aligned. 
Further, BERTScore assigns high similarity to tokens with the same meaning, thus being robust to synonymy but oblivious to their complexity.
It is also insensitive to simplified sentence structures (e.g.~word reordering, sentence splitting).
For these reasons, BERTScore is not suited for measuring simplicity, even when several references are available.

Nonetheless, for the same exact reasons, BERTScore can effectively be used as a similarity metric for the short answers generated in \questeval, see Table~\ref{table:example_questeval_questions}.

\section{Experiments}
\label{sec:experiments}

\subsection{Implementation Details}

We report Pearson correlations obtained via the SciPy python library \cite{2020SciPy-NMeth}. For each example in the dataset, we average the score for a given dimension from all the 30 annotators.  
We use the EASSE simplification evaluation package \cite{alva-manchego-etal-2019-easse} for BLEU, SARI, and FKGL.
For BERTScore, we used the official implementation.\footnote{\url{https://github.com/Tiiiger/bert_score}}
Finally, we adapted the original \questeval implementation to support BERTScore as a similarity function to compare generated answers. We make the code available here.\footnote{\url{https://github.com/recitalAI/QuestEval/\#text-simplification}}
We use the \asset gold simplifications for metrics that require references.

\subsection{Results and Discussion}

\paragraph{Metrics Correlations on Systems Simplifications}

In Table~\ref{table:correlations}, we report the Pearson correlations for 5 evaluations metrics compared across the three dimensions. 

Both SARI and FKGL do not perform well, with low correlations (<30) on all dimensions. 
Conversely, BERTScore and \questeval obtain the highest correlations, with an edge for BERTScore on Fluency and Simplicity and \questeval leading in Meaning Preservation. 
The most surprising result is the relatively high coefficients for \questeval and BERTScore on Simplicity ($\sim$40): BERTScore is robust to synonyms, so it is not really adapted to evaluating lexical simplification, and \questeval only evaluates sentences relative to their factual content, regardless of their complexity.
Therefore, neither should be equipped to measure Simplicity.
In contrast with previous findings \cite{xu-etal-2016-optimizing}, we also observe BLEU outperforming SARI on Simplicity. 

Also visible in Table~\ref{table:correlations} are the strong inter-correlations between the three evaluated dimensions: e.g.~the Fluency correlates with the Meaning better than any metric (79.1 Pearson coefficient). 
These inter-correlations between the evaluated dimensions could result in undesired spurious correlations for the metrics. This would explain why BERTScore, \questeval, or BLEU, correlate better for Simplicity than SARI or FKGL.

\paragraph{Right for the wrong reasons?}

In order to get a deeper understanding of the real correlations of the metrics, one needs to limit the inter-correlations between the different dimensions. With this purpose in mind, we compute the correlations this time on \emph{Human-Likert}, our corpus of human annotation on human written simplifications instead of system generated ones. We report the results in the right half of Table~\ref{table:correlations}. 

All inter-correlations are lower than for system-generated simplifications although still high. In particular, the Meaning is less impacted by the Fluency and the Simplicity. This allows a clearer analysis of the intrinsic metrics correlations, leading to very different conclusions.

It is probably much more difficult for the annotator to score differently human level outputs than systems outputs. This is confirmed by the lower inter-agreement obtained on \emph{Human-Likert} than on \emph{System-Likert}, while both were constituted following the exact same methodology. Therefore, it is natural that the correlations are lower for all the metrics.

With respect to Simplicity, we observe as expected that FKGL performs the best, with the only significant result (34.7).

For Meaning Preservation, \questeval achieves the highest and only significant correlation. This result is emphasized by - this time - the slight anti-correlation on Simplicity and Fluency. 
We also observe that BERTScore correlates slightly on all the dimensions but with no statistical significance.
These results confirm that inter-correlations between dimensions cause spurious correlations among automatic metrics, when evaluated on system generated simplifications.
In other words, a system generating meaning preserving or fluent \emph{but not simpler} texts, might score higher with \questeval or BERTScore. \emph{This does not necessarily mean that the system is better for Sentence Simplification!}  

Finally, it is interesting to note that on highly fluent simplifications, i.e.~human written, FKGL measures Simplicity well, while \questeval measures effectively Meaning Preservation. As both metrics are reference-less, their usage might be appropriate on any corpus even in absence of gold-references.

\section{Conclusion}

In this paper we adapted a Question-based metric to Sentence Simplification. By using BERTScore for the similarity function, we provide a smoother way to compare two answers than in the original metric, allowing to take into account synonyms. 

Further, we conducted an extensive analysis of the metrics for Sentence Simplification on both system-generated and human-written examples. On system-generated simplifications, we show that both BERTScore and \questeval improve over BLEU and SARI. 
However, on the human-written simplifications, we raise concerns about very low correlations for most of traditional metrics: actually only FKGL and \questeval are able to significantly measure Simplicity and Meaning Preservation. This paper thus calls for more frequent re-evaluation of the metrics, along with the systems advances. 

In future work, we aim at study how the combination of those metrics could constitute a global metric to evaluate Sentence Simplification.

\bibliography{anthology,custom,simplification_bibliography}
\bibliographystyle{acl_natbib}

\clearpage
\appendix
\section*{Appendix}

\section{Human-Likert Collection Methodology} \label{collection_methodology}
We follow the methodology of \cite{alva2020asset} and reuse the same interface provided by the authors.
We collect annotations using Amazon Mechanical (AMT).
The requirements for annotators are exactly the same as \cite{alva2020asset}, namely: (1) have a HIT approval rate $>= 95\%$; (2) have a number of HITs approved $> 1000$; (3) are residents of the United States of America, the United Kingdom or Canada; and (4) passed the corresponding Qualification Test designed for by the authors and provided on their repository.

The qualification test consists in a training session explaining what is sentence simplification and a rating session where the annotators had to rate 6 pairs of source-simplification pairs.
Annotators were asked to use a (0: Strongly disagree - 100: Strongly agree) continuous scale to rate sentences on three aspects represented by the following statements:
\begin{enumerate}
    \item The Simplified sentence adequately expresses the meaning of the Original, perhaps omitting the least important information.
    \item The Simplified sentence is fluent, there are no grammatical errors.
    \item The Simplified sentence is easier to understand than the Original sentence.
\end{enumerate}

The 6 sentence pairs evaluated are the same as in \cite{alva2020asset}, and were chosen to represent various simplification operations and typical errors in meaning preservation or fluency.
We then manually evaluated qualification tests to filter out spammers or workers that didn't perform the task correctly.

We used the same 100 source sentences as the System-Likert corpus and sampled one simplification each from either \asset, \turkcorpus, or \hsplit, resulting in 100 unique source-simplification pairs.
We finally collected 30 ratings per pair and per dimension (fluency, meaning, simplicity) resulting in 9000 total ratings.

\end{document}